\documentclass{article}
\usepackage{naturetex}
\usepackage{endfloat}

\begin{document}

\maketitle

{\setstretch{1.0}
	\section*{Abstract}
The fields of therapeutic application and drug research and development (R\&D) both face substantial challenges, i.e., the therapeutic domain calls for more treatment alternatives, while numerous promising pre-clinical drugs have failed in clinical trials. One of the reasons is the inadequacy of Cross-drug Response Evaluation (CRE) during the late stages of drug R\&D. Although in-silico CRE models bring a promising solution, existing methodologies are restricted to early stages of drug R\&D, such as target and cell-line levels, offering limited improvement to clinical success rates. Herein, we introduce DeepCRE, a pioneering AI model designed to predict CRE effectively in the late stages of drug R\&D. DeepCRE outperforms the existing best models by achieving an average performance improvement of 17.7\% in patient-level CRE, and a 5-fold increase in indication-level CRE, facilitating more accurate personalized treatment predictions and better pharmaceutical value assessment for indications, respectively. Furthermore, DeepCRE has identified a set of six drug candidates that show significantly greater effectiveness than a comparator set of two approved drugs in 5/8 colorectal cancer organoids. This demonstrates the capability of DeepCRE to systematically uncover a spectrum of drug candidates with enhanced therapeutic effects, highlighting its potential to transform drug R\&D.
}

\newpage
\section*{Introduction}

The fields of therapeutic application and drug research and development (R\&D) both face substantial challenges~\cite{dolgin2024future, o2023nci, wouters2020research, sean2023drug}. On the one hand, in therapeutic applications, the limited efficacy of existing treatment options is a critical concern for medical researchers~\cite{o2023nci}. This challenge is highlighted in a recent article in Nature, entitled "The future of precision cancer therapy might be to try everything", advocating for a comprehensive exploration of treatment alternatives~\cite{dolgin2024future}.  Unfortunately, the reality is characterized by a constricted repertoire of available drugs~\cite{o2023nci}. On the other hand, the domain of drug R\&D confronts the phenomenon known as "Eroom's Law", which posits that, counterintuitively, the efficiency of developing new drugs is declining despite advancements in technology~\cite{scannell2012diagnosing}. The failure of numerous promising pre-clinical drugs during clinical trials highlights the obstacles to expanding the spectrum of effective treatments.

\noindent A pivotal factor exacerbating these challenges is the inadequacy of Cross-drug Response Evaluation (CRE). CRE enables the identification of the most effective drugs at various stages of drug R\&D, which contributes to reducing costs and improving efficiency during the drug development stage~\cite{ringel2020breaking}, as well as in improving prognosis outcomes during the therapeutic stage~\cite{o2023nci, kornauth2022functional, snijder2017image}. However, due to technological and ethical constraints, CRE has largely been restricted to the target and cell line levels~\cite{luo2024toward, kuenzi2020predicting}, which represent the early stage of drug development (Fig. 1a). Hopefully, Artificial Intelligence (AI) has emerged as a key technology with the potential to predict the CRE in the late stage (Fig. 1a), offering a new approach to exploring treatment strategies more effectively~\cite{honkala2022harnessing, vijayan2022enhancing}.

\noindent Recently, a line of research have been proposed to develop in-silico CRE models~\cite{dincer2020adversarial, he2022context, kuenzi2020predicting, cadow2020paccmann, zhu2022tgsa}, while they all come with some limitations. Single-Drug Learning (SDL) models, such as ADAE~\cite{dincer2020adversarial} and Code-AE~\cite{he2022context}, have demonstrated potential in developing personalized treatment and identifying patient biomarkers. Nonetheless, the SDL paradigm, employing a independent model for each specific drug, lacks CRE capabilities due to difficulties in comparing results across different models (Supplementary Fig. 1a). This limitation hampers their ability to identify the most effective drug for specific patients and indications, thereby restricting their broader applications. In contrast, Multi-Drug Learning (MDL) models, including DrugCell~\cite{kuenzi2020predicting}, Paccmann~\cite{cadow2020paccmann}, and TGSA~\cite{zhu2022tgsa}, leverage extensive drug response data and facilitate cross-drug response comparisons~\cite{chen2021survey}. However, these MDL models are generally restricted to the cell-line level~\cite{chen2021survey}, which may result in suboptimal performance when predicting patient-level drug responses due to the disparities between cell lines and patients.

\noindent In this paper, we introduce DeepCRE, a pioneering AI model designed to predict CRE effectively in the late stages of drug R\&D. This approach aims to tackle the substantial challenge of the limited effective treatment options in both therapeutic applications and drug R\&D. We demonstrate the utility of DeepCRE through patient- and indication-level CRE analyses, highlighting its unique advantages for therapeutic applications. To further substantiate the transforming potential of DeepCRE in drug R\&D, we employ wet lab assays, which validate its superior performance. Specifically, DeepCRE outperforms the existing state-of-the-art (SOTA) models~\cite{dincer2020adversarial, he2022context, kuenzi2020predicting, cadow2020paccmann, zhu2022tgsa} by achieving an average performance improvement of 17.7\% in patient-level CRE, and a 5-fold increase in indication-level CRE. These advancements enable more precise predictions of personalized treatments and offer improved pharmaceutical value assessment for specific indications, respectively. Furthermore, DeepCRE has identified a set of six drug candidates that show significantly greater effectiveness than a comparator set of two approved drugs in 5/8 colorectal cancer organoids. To note, it is not merely about identifying one or two potential drugs, rather, it highlights the capability of DeepCRE to systematically uncover a spectrum of drug candidates with enhanced therapeutic effects. 

\newpage
\section*{Results}\label{results}

\subsection*{Overview of DeepCRE}
The drug research and design (R\&D) process encompasses target identification, cell line assays, animal or organoid testing, and clinical trials~\cite{kaitin2010deconstructing}. A significant number of drug candidates fail at various stages, resulting in inefficient drug development~\cite{scannell2012diagnosing, dickson2009cost}. Interestingly, there is a magical connection between the success rate and the CRE abundance in every stage of the drug R\&D, where multiple CRE exist in the early stage while few CRE remains in the late stage (Fig. 1a). Enhancing the presence of CREs in later stages, especially with the aid of in-silico CRE models, could markedly improve drug development outcomes (Fig. 1a,b).

\noindent The principle of DeepCRE is to project the cell line and patient data onto a shared space (Fig. 1c). This approach enables the use of a drug response prediction model, trained using cell line data, to accurately predict patient drug responses. This methodology has contributed to an average performance boost of 17.7\% and a 5-fold improvement in patient and indication-level CRE prediction, respectively, surpassing existing SOTA models (Fig. 1c). Notably, DeepCRE has identified six drug candidates (Set A) that exhibit significantly higher effectiveness than two approved drugs (Set C) in 5 out of 8 CRC organoids tested (Fig. 1c-e). To note, it is not merely about identifying one or two potential drugs, rather, it highlights the capability of DeepCRE to uncover a collection of drug candidates with enhanced therapeutic effects.

\begin{figure}[h]
	\begin{center}
		\includegraphics[width=0.75\textwidth]{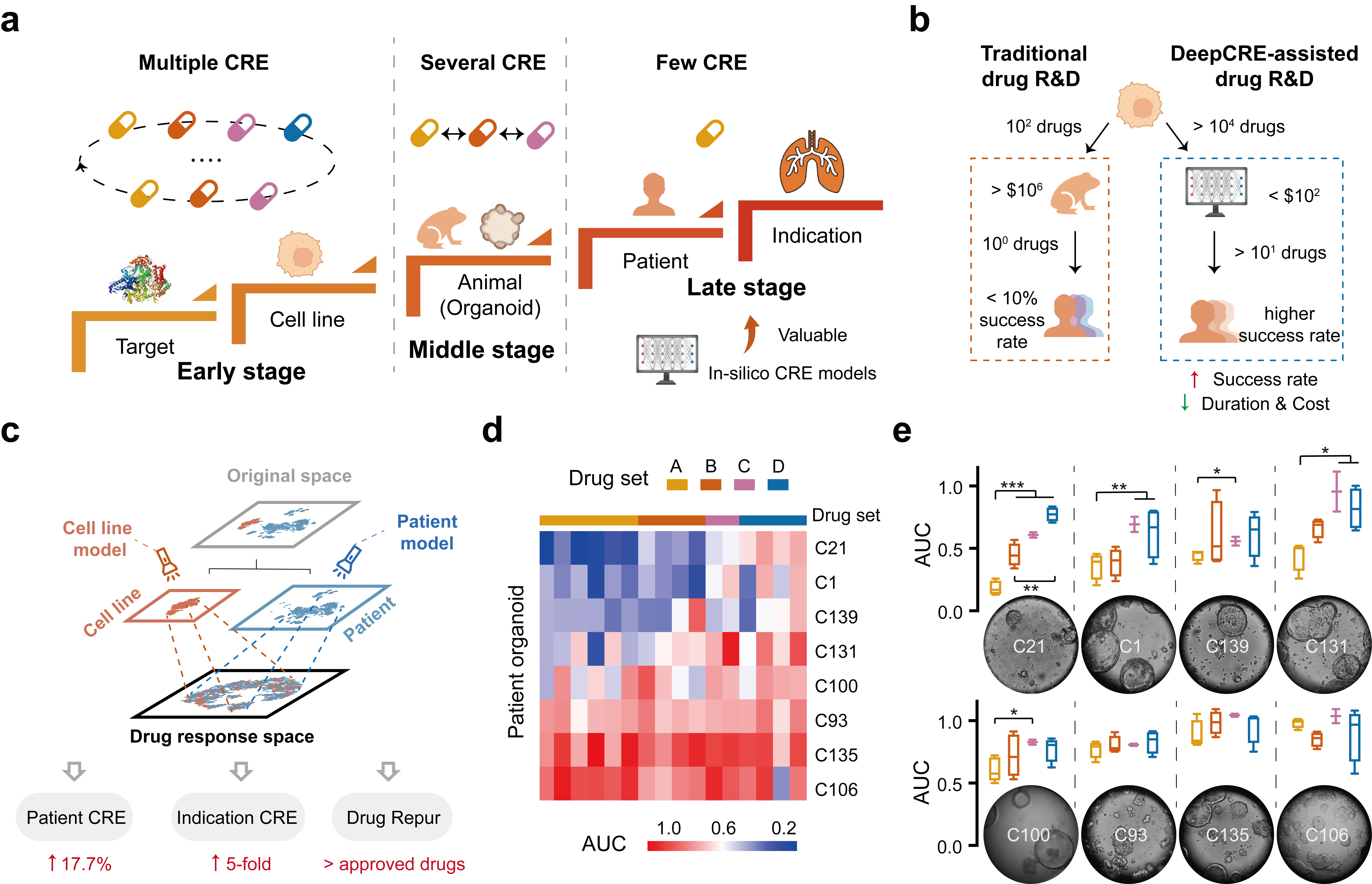}
	\end{center}
        \captionsetup{labelfont=bf}
        \caption{\textbf{The objective and application of DeepCRE.} \\
        \textbf{a,} Illustration of CRE distribution across different drug development stages, highlighting the potential benefits of in-silico CRE models in enhancing CRE availability during late-stage development.
         \textbf{b,} Comparative diagram showing the traditional vs. DeepCRE-assisted drug development processes. The DeepCRE-assisted method tends to increase success rates while reducing both time and costs.
        \textbf{c,} Conceptual framework of DeepCRE, aligning cell line and patient representations with the drug response space. This alignment enables the extrapolation of cell line drug responses to predict patient drug responses effectively. 
        \textbf{d,} Heatmap displaying the Area Under the Curve (AUC) scores for eight patient-derived colorectal cancer (CRC) organoids treated with four different sets of drugs.
        \textbf{e,} Boxplot illustrating the AUC scores of the four drug sets, with Set A (identified by DeepCRE) demonstrating significantly greater efficacy than Set C in 5/8 CRC organoids.
}
	\label{fig:fig1}
\end{figure}

\subsection*{The construction of the DeepCRE model}
We constructed the first patient-level CRE dataset and established the initial DeepCRE model (Supplementary Fig. 1-3 and Supplementary Table 1). Upon identifying the limitations of the initial DeepCRE model (details refer to Supplementary materials, Supplementary Fig. 4a,5), we turned to the architecture of domain separation network (DSN), which is a strategy for transfer learning and has been successfully applied in computer vision~\cite{ganin2016domain, bousmalis2016domain, csurka2017domain}. We propose the DeepCRE model zoo, comprising a suite of the DSN architecture (Fig. 2a) and its derivatives (Fig. 2b). The DeepCRE model zoo achieves an average performance increase of 17.21\%, 11.52\% and 7.69\% over P-SDL models, C-MDL models and their combined counterparts, respectively (Supplementary Fig. 4b). Nonetheless, there remain four tumor types in which the DeepCRE model zoo demonstrates only average performance (Supplementary Fig. 5). Additionally, we observe no distinct differences in performance among these DeepCRE models (Supplementary Table 2). Since these models primarily differ in pretraining stages, we hypothesize that some pretraining settings may not be suitable for specific tumor types.

\noindent Diving deeper into the transferability difference of the pretraining data (details refer to Supplementary materials, Supplementary Fig. 6 and Supplementary Table 3,4), we propose a selective pretraining strategy for the patient unlabeled data. Contrary to the generic all-data pretraining method, our tumor type-adaptive pretraining strategy specifically aligns T2 patients (replace all patients) with all cell lines during the pretraining stage to enhance the accuracy of drug response predictions for T2 patients (Fig. 2c).

\noindent Employing this adaptive pretraining strategy significantly enhances the performance of our adv-loss-based DeepCRE models, contributing to the average performance increases of 3.51\%, 4.92\%, and 11.28\% for AE-adv, DSRN-adv, and DSN-adv models, respectively (Fig. 2d and Supplementary Table 5). The DSN-adv model, in particular, stands out for its outstanding performance across various tumor types (Supplementary Fig. 7a and Supplementary Table 6). This leap in performance underscores the pivotal role of our pretraining strategy, especially in the alignment of cell line and patient gene expression profiles (GEPs). A comprehensive comparative analysis detailing these pretraining strategies is available in the Supplementary materials (Supplementary Fig. 7b,c).

\begin{figure}[h]
	\begin{center}
		\includegraphics[width=0.75\textwidth]{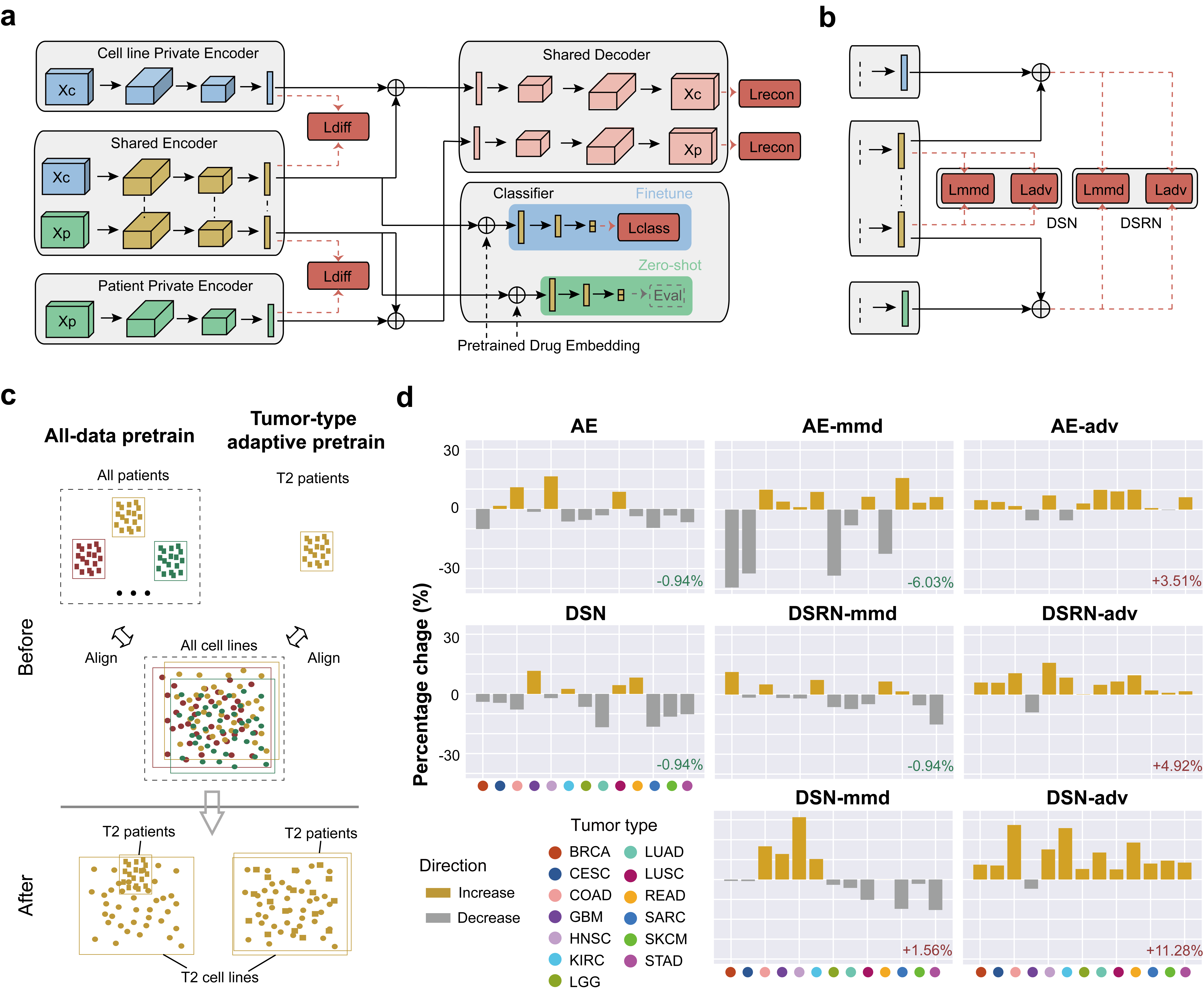}
	\end{center}
        \captionsetup{labelfont=bf}
        \caption{\textbf{The construction of the DeepCRE models.} \\
        \textbf{a,} Illustration of DeepCRE: the representations of cell lines and patients are aligned using a domain separation network (DSN)~\cite{bousmalis2016domain} architecture.
        \textbf{b,} Variants of the DeepCRE model. In the DSN setting, the private embeddings of the source domain and target domain can be kept similar via optimizing $L_{mmd}$ (DSN-mmd) or $L_{adv}$ (DSN-adv). Moreover, another variation is that the concatenation of private and shared embeddings is kept similar, which can be represented as DSRN-mmd and DSRN-adv.
        \textbf{c,} Two pretraining strategies lead to different alignments of cell lines and patients of one specific tumor type.
        \textbf{d,} Bar plots showing the percentage change in performance from all-data pretraining to tumor-type adaptive pretraining for eight DeepCRE models across 13 tumor types, with the numbers below representing the average percentage change among all tumor types. Three adv-loss-based DeepCRE models (the last column) generally improve significantly in 12 tumor types, except for GBM. 
 }
	\label{fig:fig2}
\end{figure}

\subsection*{DeepCRE demonstrates superior performance in patient-level CRE across 13 tumor types}
The DeepCRE models, benefiting from an enhanced pretraining strategy, exhibit SOTA performance across all tumor types, surpassing both P-SDL and C-MDL models (Fig.3a and Supplementary Table 6). On average, DeepCRE models demonstrate performance improvements of 28.21\%, 22.07\%, and 17.77\% over P-SDL models, C-MDL models, and their combined counterparts, respectively (Supplementary Fig. 4c). Notably, the DSN-adv model emerges as a standout performer, achieving SOTA results in 12 out of 13 tumor types (Fig. 3a), and displaying average performance increases of 27.49\%, 21.38\%, and 17.08\% over P-SDL models, C-MDL models, and their combined counterparts, respectively (Supplementary Fig. 4d). Generally, the AUROC performance of DeepCRE increases from the initial AE model to the DSN, DSN-mmd, and DSN-adv models.

\noindent To delve into the underlying mechanisms of this performance boost, we quantify the alignment effect (after pretraining) using two metrics: the relative maximum mean discrepancy~\cite{{gretton2012kernel}} (MMD) and Kullback-Leibler (KL) divergence~\cite{kullback1951information}. Both metrics demonstrate a decreasing trend from the original GEPs to the AE, DSN, DSN-mmd, and DSN-adv encoded embeddings, indicating enhanced alignment across these pretraining methods (Fig. 3b,c and Supplementary Table 7). Four tumor types exhibit a relative MMD of less than 0.3 for the DSN-adv model, as denoted by an asterisk (*) in Fig. 3b, and are further elucidated in the t-SNE plots (Fig. 3d).

\noindent Furthermore, the DeepCRE DSN-adv model outperforms all of the P-SDL and C-MDL models across all tumor types (Fig. 3e). To explore the association between the model performance and paradigm selection, we compare the performance of DSN-adv with the best P-SDL and C-MDL models, respectively (Supplementary Fig. 8a and Supplementary Table 8). The percentage increase in performance of the DSN-adv model compared to the best P-SDL (brown) and C-MDL (blue) models illustrates the variability of pan-cancer improvement (Fig. 3f). The greatest increase for the best P-SDL model is observed in SARC, while the maximum increase for the best C-MDL model is seen in READ (Fig. 3f and Supplementary Fig. 8b). Notably, the tumor type COAD, which is associated with READ, also displays a substantial increase (Fig. 3f and Supplementary Fig. 8b). Intriguingly, a strong negative correlation (r = 0.782) is observed between the drug overlap percentage and the log-scaled increase percentage for the P-SDL model (Supplementary Fig 8c,d and Supplementary Table 8). This suggests that the SDL paradigm, which lacks drug encodings, exhibits subpar performance when predicting unseen drugs~\cite{chen2021survey}. In contrast, for the C-MDL mode, mutation-data-based methods generally outperform expression-data-based methods (Fig. 3g). We speculate that this is due to two reasons: i) transfer learning (alignment) is less crucial for mutation-based methods, as mutations primarily occur in tumor cells~\cite{martincorena2015somatic}; ii) transfer learning is essential for expression-based methods, as bulk expression data comprises not only tumor cells but also immune and stromal cells~\cite{aran2015systematic}.

\begin{figure}[h]
	\begin{center}
		\includegraphics[width=0.75\textwidth]{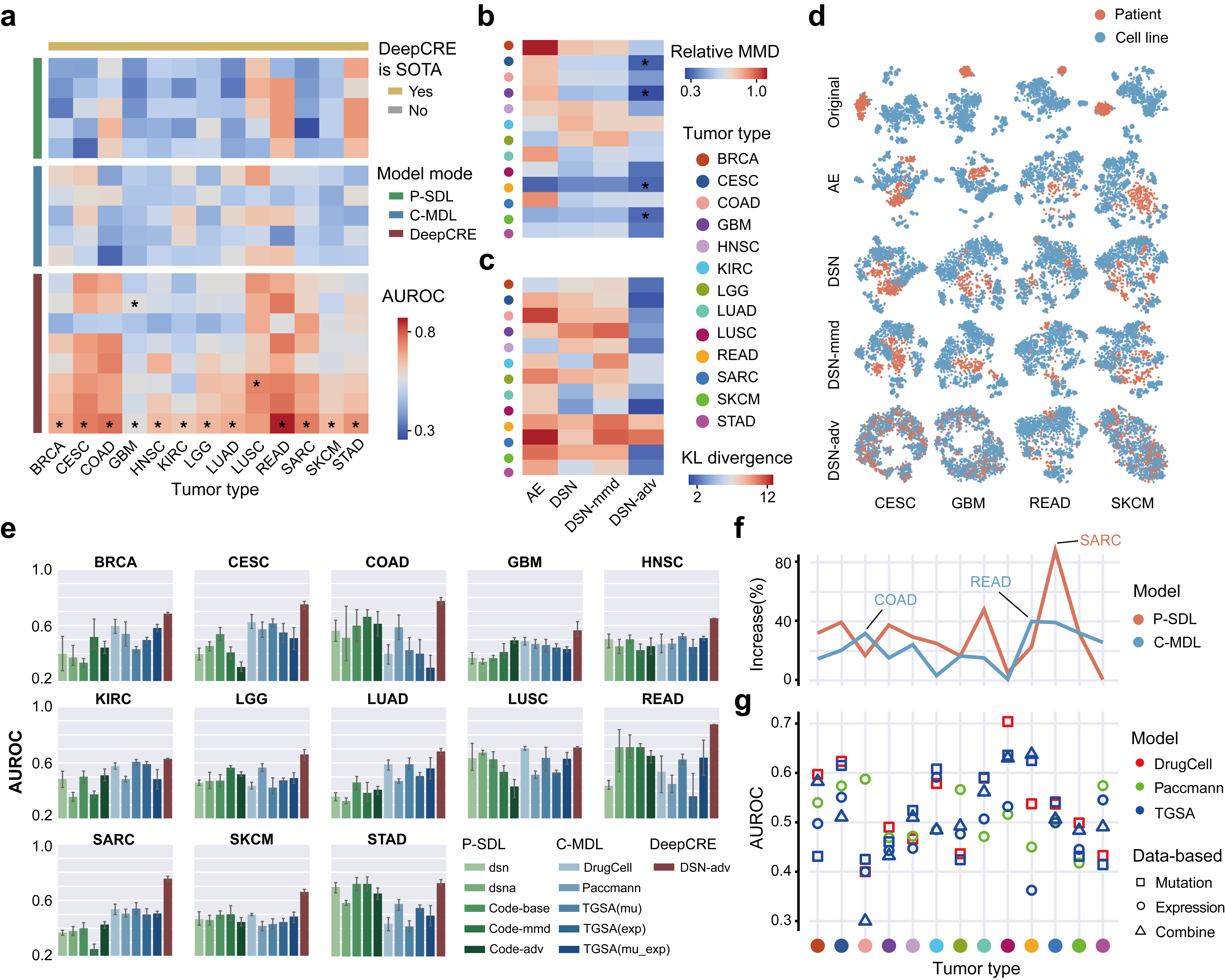}
	\end{center}
        \captionsetup{labelfont=bf}
        \caption{\textbf{DeepCRE outperforms other models in patient-level CRE across 13 tumor types.}\\
        \textbf{a,} Heatmap comparing the AUROC performance of P-SDL, C-MDL, and tumor-type adaptive pretraining DeepCRE models. The listed models (from top to bottom) are  P-SDL (dsn, dsna, Code-base, Code-mmd, Code-adv), C-MDL (DrugCell, Paccmann, three types of TGSA) and DeepCRE (AE, DSN, AE-mmd, DSRN-mmd, DSN-mmd, AE-adv, DSRN-adv, DSN-adv) models. DeepCRE models achieve SOTA (marked with *) performance in all tumor types.
        \textbf{b,c,} Heatmaps displaying alignment metrics for patients (brown) and cell lines (blue) across four pretraining methods: \textbf{b,} Maximum Mean Discrepancy~\cite{gretton2012kernel} (MMD) relative to original expression within each tumor type. The relative MMD generally decreases from AE, DSN, DSN-mmd to DSN-adv, indicating improved alignment. Tumor types with relative MMD of DSN-adv less than 0.3 are marked with * for further analysis in \textbf{d}. \textbf{c,} Kullback-Leibler (KL) divergence~\cite{kullback1951information} also decreases from AE, DSN, DSN-mmd to DSN-adv. The KL divergence of the original expression is not shown due to its nearly infinite value.
        \textbf{d,} t-SNE plots of gene expression encoded embeddings for four tumor types (marked with * in \textbf{b}) from original expression to four DeepCRE models. Notably, DSN-adv (the last row) exhibits significant improvement in the alignment of patients and cell lines compared with other models.
        \textbf{e,} Bar plots indicating the AUROC performance of P-SDL, C-MDL and the DeepCRE DSN-adv model. DSN-adv outperforms all P-SDL and C-MDL models across all tumor types.
        \textbf{f,} Line chart showing the percentage increase in performance of the DSN-adv model compared with the best P-SDL (brown) and C-MDL (blue) models across all tumor types. The maximum increase for the best P-SDL model is observed in SARC, while the maximum increase for the best C-MDL model is seen in READ. The READ-related tumor type COAD also exhibits a substantial increase.
        \textbf{g,} Scatter plot indicating the AUROC performance of C-MDL models, with different shapes representing the types of used data. Generally, mutation-data-based methods outperform expression-data-based methods for C-MDL models.
}
	\label{fig:fig3}
\end{figure}

\subsection*{DeepCRE's potential in clinical pharmaceutical value assessment for pre-clinical drug candidates}
To assess DeepCRE's potential in clinical pharmaceutical value assessment for pre-clinical drug candidates, we have evaluated 233 small molecules across 13 tumor types (indications) and verified them through the DrugBank~\cite{wishart2018drugbank}, ClinicalTrials~\cite{zarin2011clinicaltrials}, and Repurposing Hub~\cite{corsello2017drug} databases (Fig. 4a-c and Supplementary Fig. 9).

\noindent The pharmaceutical value assessment results are initially validated using the DrugBank database, which contains approved or investigational drugs with high confidence. Among the 17 small molecules listed in the DEI table, 10 drugs were recorded in DrugBank, while seven small molecules lack any record (Fig. 4d and Supplementary Table 10). Visualized as a heatmap, the DEI table highlights predicted-efficient EDIs in purple (and DrugBank-recorded EDIs in blue), while predicted-inefficient EDIs are shaded grey (Fig. 4e). Notably, all DrugBank-recorded EDIs are predicted-efficient, indicating no false negatives. Furthermore, purple-colored EDIs represent "promising" opportunities for new drug development.

\noindent To further validate the predictions, we cross-reference in-clinical-test records from the ClinicalTrials database (Supplementary Table 11). Drug candidates identified by DeepCRE exhibit anti-tumor evidence in terms of mechanism of action (MoA) and current indications (Supplementary Fig. 10a,b). Nine drug candidates are in clinical testing, and after excluding widely-used Gemcitabine (Supplementary Fig. 10c,d), they are verified to be tested in various indications, including the 13 types of tumor indications used in our study (Fig. 6f and Supplementary Table 12). Mirroring the DrugBank analysis, the DEI table is visualized using a heatmap, which also includes additional information about the tumor types in which the drug candidates were undergoing clinical testing (Fig. 6g). The predicted results demonstrate excellent agreement with the in-clinical-test records, indicating strong concordance between DeepCRE's predictions and real-world clinical data (details refer to the Supplementary materials). Furthermore, five drug candidates are identified as Qualified drugs by DeepCRE (Fig. 6g), while no more than one Qualified drug candidate is discovered by any previous SOTA models (Supplementary Fig. 11). Additionally, the EDIs and Qualified drugs identified by our model are associated with 115 and 105 in-clinical-test records, respectively, representing substantial increases of 117\% and 400\%, respectively, compared to previous SOTA models (Fig. 6h,i). This result means the drug candidates discovered by our model show at least a 5-fold increase in indication-level CRE compared to any previous SOTA models.

\noindent Finally, we validate the DEI table through the Repurposing Hub database (Supplementary Fig. 12), which collects information on existing therapeutics repurposed for new disease indications from published papers~\cite{corsello2017drug}. Overall, the predictions from the DeepCRE are largely consistent with fully validated, in-clinical-test, and existing experimental evidence, indicating the strong alignment of our model with the expertise of real-world professionals in assessing the pharmaceutical value for pre-clinical drug candidates.

\begin{figure}[h]
	\begin{center}
		\includegraphics[width=0.75\textwidth]{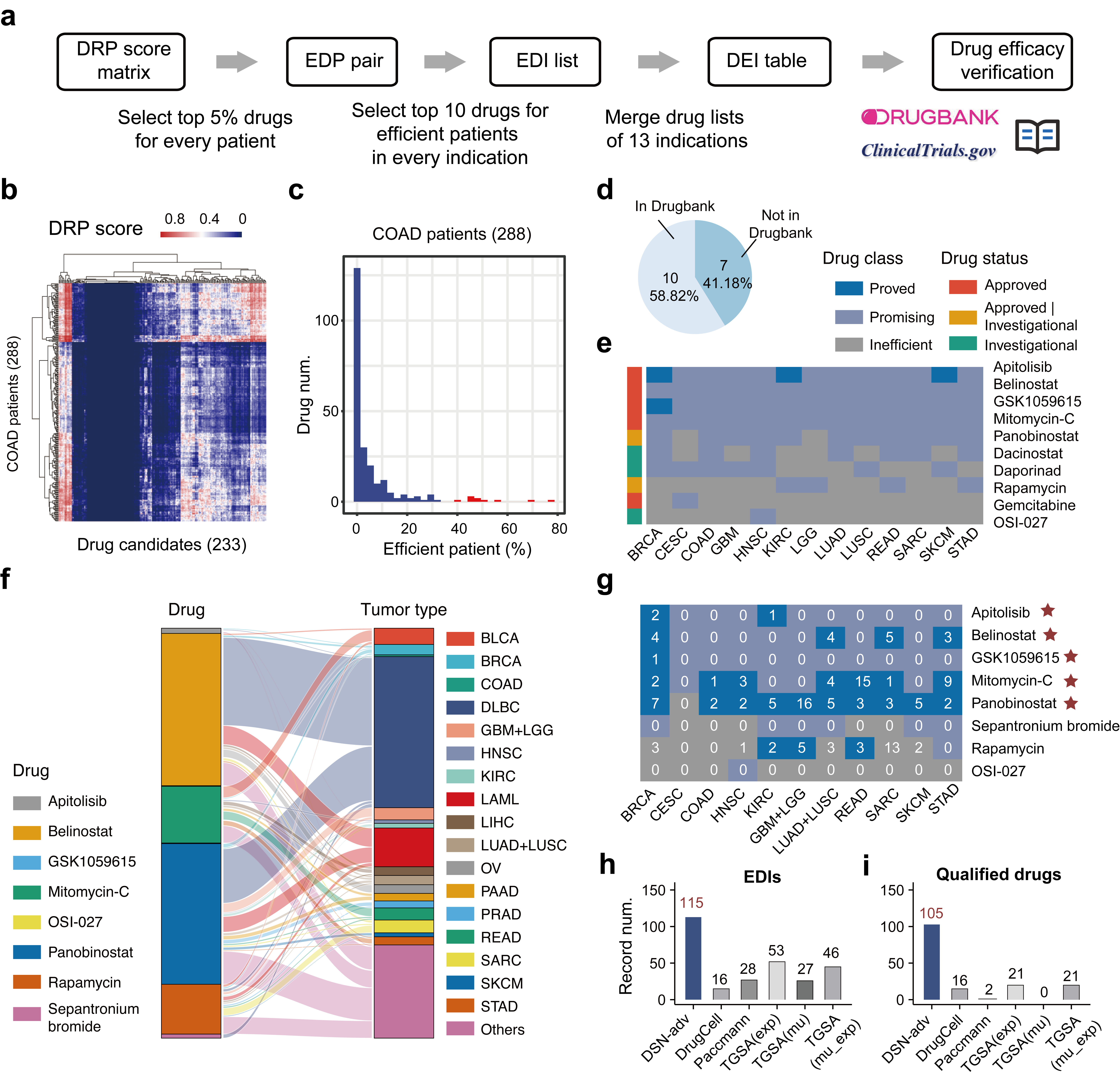}
	\end{center}
        \captionsetup{labelfont=bf}
        \caption{\textbf{DeepCRE outperforms other models in indication-level CRE.}\\
        \textbf{a,} Workflow illustrating the validation process of drug candidates discovered by DeepCRE. Refer to the Methods section for a detailed description. DRP: drug response for patients. EDP: efficient drugs for patients. EDI: efficient drugs for indications. DEI: drug efficacy for indications.
        \textbf{b,} Heatmap displaying DRP score matrix, e.g., COAD, comprising 288 patients (rows) and 233 drug candidates (columns).
        \textbf{c,} Histogram plot of EDPs for COAD. The top 10 drug candidates with the highest efficiency in patients are highlighted in red. Three drug candidates are efficient for over 50\% of COAD patients.
        \textbf{d,} Pie chart demonstrating 10 out of 17 drug candidates have records in Drugbank.
        \textbf{e,} Heatmap of the DEI table. Both predicted-efficient and Drugbank-recorded EDIs are colored blue and labeled “Proved”. Only predicted-efficient EDIs are colored purple and labeled “Promising”. Predicted-inefficient EDIs are colored grey and labeled “Inefficient”.
        \textbf{f,} Sankey plot depicting the distribution of in-clinical-test tumor types for the drug candidates.
        \textbf{g,} Heatmap of the in-clinical-test records, colored by the combination of predicted results and in-clinical-test records. Qualified drugs are labeled with a pentagram. 
        \textbf{h,i,} Bar plots demonstrating the record numbers of the DSN-adv method and five C-MDL methods: \textbf{h} for EDIs, \textbf{i} for Qualified drugs.
}
	\label{fig:fig4}
\end{figure}

\subsection*{Validation of DeepCRE for efficient drug candidates identification in the CRC21 patient}
DeepCRE's capability to identify efficient drug candidates is further validated in a clinical setting, focusing on the CRC21 patient. Despite receiving XEOLX treatment, the patient experienced relapse after 13 days. Tumor samples collected before treatment are used to construct CRC organoids, enabling the evaluation of drug candidates identified by DeepCRE (Fig. 5a). This In-SMARchip drug testing methodology has been previously validated in lung and colorectal tumor organoids~\cite{hu2021lung, wu2022grouped, wang2018nanoliter}. The drug candidates are categorized into four sets based on their MoAs and the evaluation variance between DeepCRE and traditional methods (Fig. 5b,c and Supplementary Fig. 13). We try to keep the MoAs diversity in each drug set (Fig. 5b) and ensure the drug set with the largest evaluation disparity for comparison (Fig. 5c). 

\noindent The drug testing results prove that DeepCRE outperforms traditional methods in two aspects. Firstly, six drug candidates (Set A) identified by DeepCRE exhibit significantly greater efficacy than a comparator set of two approved drugs (Set C), one of which is "Oxaliplatin combined with 5-Fluorouracil", the XEOLX treatment itself (Fig. 5d). Secondly, drug candidates with specific MoAs, such as PI3K/mTOR signaling and Chromatin histone acetylation, show potential to exceed traditional chemotherapy (e.g., drugs with the MoA of DNA replication) treatments (Fig. 5e).

\begin{figure}[h]
	\begin{center}
		\includegraphics[width=0.75\textwidth]{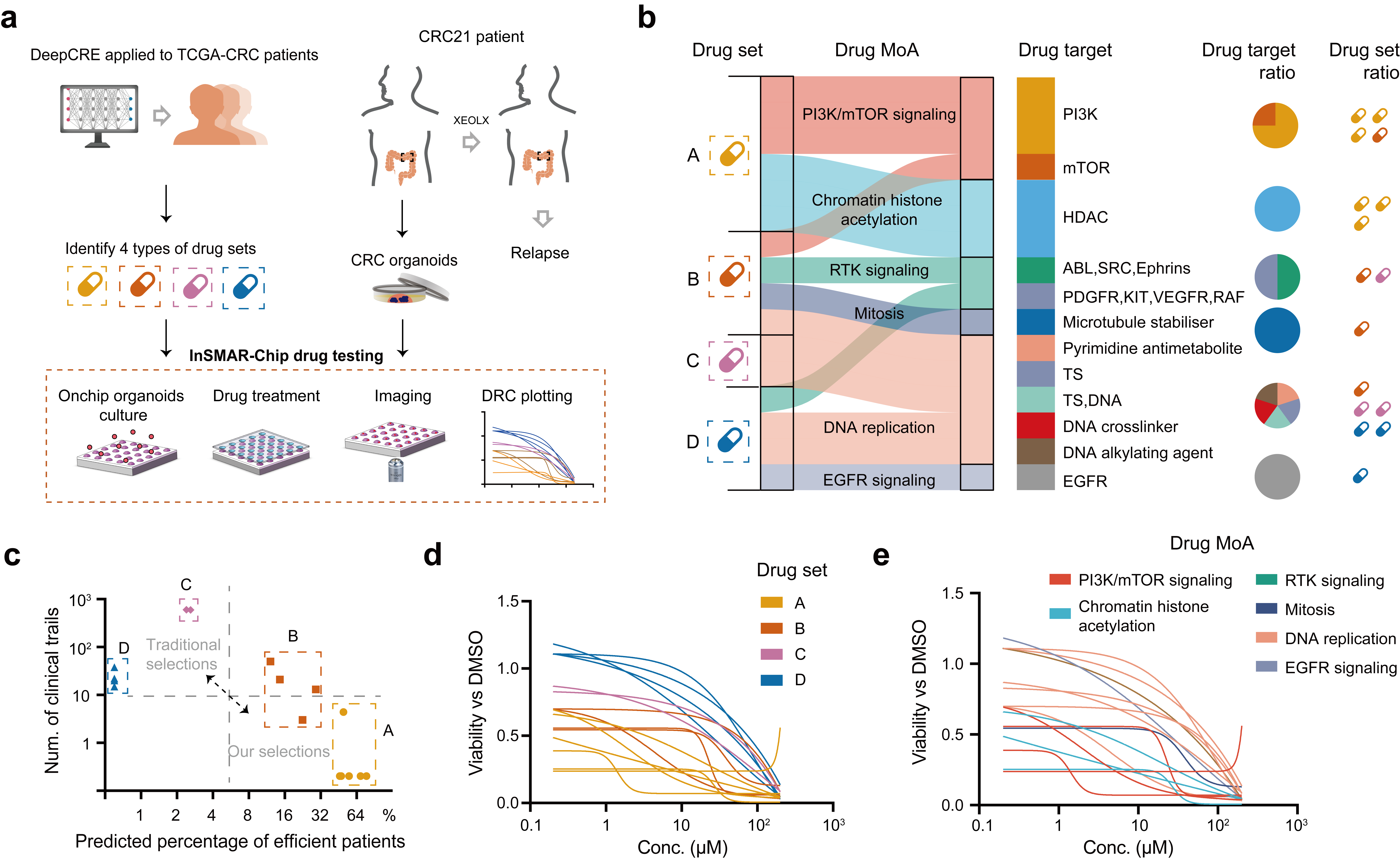}
	\end{center}
        \captionsetup{labelfont=bf}
        \caption{\textbf{Validation of DeepCRE for efficient drug identification in the CRC21 patient.} \\
        \textbf{a,} Workflow demonstrating the utilization of DeepCRE to identify four types of drug sets, which were subsequently evaluated using CRC organoids derived from a patient who relapsed after traditional XEOLX treatment, employing InSMAR-Chip drug testing methodology. DRC: drug response curve.
         \textbf{b,} Sankey plot illustrating four drug sets with diverse mechanisms of actions (MoAs) and targets.
        \textbf{c,} 2D XY-plot displaying the evaluation outcomes of DeepCRE as well as traditional methods.
        \textbf{d,} DRC colored by the drug set indicating the ranking of drug efficacy: A\textgreater B\textgreater C\textgreater D.
        \textbf{e,} DRC colored by the drug MoA indicating that certain drugs with specific MoAs exhibit relatively greater efficacy, such as PI3K/mTOR signaling and Chromatin histone acetylation.
}
	\label{fig:fig5}
\end{figure}

\subsection*{Drug repurposing validation of DeepCRE in eight CRC organoids}
To further validate the potential of DeepCRE in drug repurposing for CRC, seven more CRC organoids are included for In-SMARchip drug testing. Among the eight CRC organoids tested, five exhibits a significant efficacy increase of set A over set C (Fig. 1d,e, Fig. 6a and Supplementary Fig. 14a). Moreover, drug candidates in set A are also more efficient than two approved drugs in set C in the other three CRC organoids (Supplementary Fig. 14b). Generally, the efficacy patterns of drug candidates were consistent across different patients, except for Sorafenib (D3), one of the receptor tyrosine kinase (RTK) inhibitors (Fig. 6b). However, there were variations in the exact correlation coefficients: inhibitors of PI3K/mTOR signaling and Chromatin histone acetylation show coefficients above 0.96, while those targeting DNA replication display coefficients ranging from 0.33 to 0.93, reflecting the diversity and complexity of chemotherapy~\cite{alam2018chemotherapy, makovec2019cisplatin} (Fig. 3c). 

\noindent Some intriguing insights emerge from the drug testing results, shedding light on the deeper mechanism exploration of drug candidates with similar MoAs. For instance, the DRC of GSK1059615 exhibits a significant increase at the highest concentration compared to other PI3K/mTOR inhibitors (Fig. 6d), except for C100 (Supplementary Fig. 14c). Perturbation data from CMAP L1000~\cite{subramanian2017next} database provides some explanations (Fig. 6e-f). Cells treated with GSK1059615 show up-regulated cellular survival and stress response related pathways (isomerase activity, intramolecular transferase activity, and phosphotransferases) and genes (IGFBP3, MYCBP, GTF2A2), DNA repair, and cytoskeleton regulation related genes (PARP1, RUVBL1, STAMBP), and down-regulated cell cycle checkpoints related pathways (protein tyrosine/MAP kinase/MAP kinase tyrosine phosphatase activity) and cell cycle related genes (STX1A, PARP6, KDM3A). The up-regulation and down-regulation of these genes and pathways may reflect the defensive mechanisms adopted by cells to resist the toxicity of high concentrations of drugs, potentially leading to the pausing of the cell cycle or the inhibition of cell death pathways~\cite{copple2010keap1, jones1986cellular}. The other example is that the DRC of the drug Sorafenib, rather than another RTK inhibitor Dasatinib, declines at higher concentrations in three CRC organoids (Fig. 6g). It has been reported that some transport proteins may play a role in the accumulation and resistance of drugs, with Dasatinib being primarily restricted by ABCG1, while Sorafenib being mainly limited by ABCG2~\cite{tang2013impact}. Further investigation suggests that ABCG2 expression, significantly lower in CRC tumor tissues compared to normal tissues~\cite{liu2018integrated} (Fig. 6h), may contribute to the reduced resistance of Sorafenib in CRC. Molecular docking analysis corroborates these findings, demonstrating a favorable binding affinity of Sorafenib to the target ABCG2 (Fig. 6i and Supplementary Fig. 14d). Similar conclusions of binding poses and interactions have been drawn by a published paper, which presents single-particle cryo-EM studies of ABCG2 binding with another RTK inhibitor imatinib~\cite{orlando2020abcg2}. Furthermore, survival analysis reveals a negative correlation between ABCG2 expression and disease-free survival (DFS), implicating ABCG2-mediated multi-drug resistance (MDR) in CRC (Fig. 6j).

\begin{figure}[h]
	\begin{center}
		\includegraphics[width=0.75\textwidth]{figures/fig6.png}
	\end{center}
        \captionsetup{labelfont=bf}
        \caption{\textbf{Drug repurposing validation of DeepCRE in eight CRC organoids.} \\
        \textbf{a,} DRC across four CRC organoids, colored by the drug set, indicating that drug set A possesses significantly greater effectiveness than drug set C.
         \textbf{b,} Heatmap of correlation coefficients of drugs, demonstrating that the drug efficacy is related to the drug MoA and the drug set.
        \textbf{c,} Boxplot of correlation coefficients of drugs, indicating differences in the distribution of efficient populations among drugs targeting DNA replication. 
        \textbf{d,} DRC of PI3K/mTOR inhibitor drugs across four CRC organoids, showing the dramatic increase in DRC of the drug GSK1059615 at the highest concentration.
        \textbf{e,} Umap plot of gene expression of cells after different perturbations, with representative differential genes shown in the dashed frame.
        \textbf{f,} Bar plot of enriched pathways, with up-regulated pathways colored in red and down-regulated pathways in blue.
        \textbf{g,} DRC of RTK inhibitor drugs across three CRC organoids, showing the decline in DRC of the drug Dasatinib, rather than Sorafenib, at higher concentrations.
        \textbf{h,} Boxplot of genes ABCG1 and ABCG2 in the TCGA-CRC cohort, indicating significantly lower expression of ABCG2 in tumor samples.
        \textbf{i,} ABCG2-Sorafenib docking (top) from the view of extracellular space of ABCG2 TM helices in the ATP bound state and (down) in 2D diagram format .
        \textbf{j,} Survival curve of ABCG2 in the disease-free survival (DFS) of the TCGA-CRC cohort.
}
	\label{fig:fig6}
\end{figure}

\newpage
\section*{Discussion}

In this study, we have proposed the DeepCRE model and demonstrated its potential to transform the drug R\&D process. Through advancements in pretraining strategies, DeepCRE surpasses the existing SOTA models, achieving an average performance improvement of 17.7\% in patient-level CRE, and a remarkable 5-fold increase in indication-level CRE. Moreover, DeepCRE has identified six drug candidates that show significantly greater effectiveness than a comparator set of two approved drugs in 5/8 CRC organoids tested. Besides, we have elucidated some pharmacodynamic and pharmacological insights into certain drug candidates, such as GSK1059615 and Dasatinib, which could inform further R\&D process. Importantly, it is not merely about identifying one or two potential drugs, rather, it highlights the capability of DeepCRE to systematically uncover a spectrum of drug candidates with enhanced therapeutic effects.

\noindent There are several future directions to explore. Firstly, leveraging Large Language Models (LLMs) for patient GEPs holds tremendous potential~\cite{zhang2024scientific, theodoris2023transfer}. By encoding GEPs with specialized genomic or multi-modal LLMs, which integrate gene languages with natural languages (clinical indicators and patient descriptions), we can derive more comprehensive insights. Such integration promises to revolutionize our understanding and approach to personalized medicine. Secondly, we propose a shift in drug R\&D from the conventional screening methodology to a generative model-based approach. Generating models enables the active design of molecules with desired properties, offering benefits such as targeted design, cost and time savings, and fostering innovation by creating novel chemical entities\cite{Rafic2023GenAI}. By moving away from the exhaustive screening of existing compounds, this strategy presents a promising avenue for the discovery of groundbreaking therapies characterized by enhanced efficacy and reduced rates of failure in clinical trials. Finally, the establishment of a synergistic dry-wet-loop framework, integrating computational predictions with experimental validations, is critical for accelerating the drug R\&D process. This harmonious integration promises a more efficient transition from pre-clinical research to clinical trials, directly translating to faster access to new treatments for patients. Such advancements underscore the importance of leveraging cutting-edge technologies and interdisciplinary approaches to overcome current limitations and fulfill the promise of next-generation medical breakthroughs.

\newpage
\section*{Methods}

\subsection*{Construction of the patient-level CRE dataset}

The patient-level CRE dataset involved the aggregation of both unlabeled and labeled data from public sources. The unlabeled dataset comprises annotations, such as tumor types and gene expression profiles sourced from the CCLE~\cite{barretina2012cancer} and UCSC Xena~\cite{goldman2018ucsc} databases. All gene expression data was standardized using the transcripts per million (TPM) metric and subsequently log-transformed to ensure uniformity in data scale and distribution. The labeled dataset integrates drug response metrics from the GDSC database~\cite{yang2012genomics} and patients from the TCGA database~\cite{liu2018integrated}. In the case of cell lines, drug response is quantified via the area under the drug response curve (AUC), offering a comprehensive measure of drug efficacy.  For patient data, the progression-free survival (PFS) period metric is utilized, reflecting the duration of response before clinical relapse~\cite{liu2018integrated}.

\noindent Data curation involved several quality control measures. Only samples with both tumor type annotations and corresponding gene expression data were considered for inclusion. Labeled data underwent further refinement, requiring the following criteria: (i) the drug was a small molecule with a readily available SMILES representation in PubChem~\cite{kim2019pubchem} or Drugbank~\cite{wishart2018drugbank}, and (ii) the treatment regimen consists of the single-drug therapy throughout the entire duration. Tumor types with sufficient amounts of unlabeled data (more than 50 samples for patients and more than 30 samples for cell lines) and labeled data (at least seven samples for patients with more than one drug overlapping in patients and cell lines) were retained for analysis (Supplementary Fig. 3). Finally, the drug response labels were binarized based on the median response within each tumor type to facilitate analysis.

\subsection*{Establishment of the DeepCRE model zoo}

The DeepCRE model zoo comprises eight models, including the initial autoencoder (AE), domain separation network (DSN), and their variants. Additional previous SOTA models, such as P-SDL and C-MDL models, are described in the Supplementary materials, along with detailed experiment setup information.

\subsubsection*{The initial autoencoder-based DeepCRE model}

As shown in Supplementary Fig. 1d, the DeepCRE models take drug and patient representations as input. Drug embeddings are generated using a pretrained Graph Neural Network~\cite{hu2019strategies} (GNN) with a hidden representation dimension of 300. Patient embeddings are generated by an autoencoder designed for the initial DeepCRE model. The autoencoder is pretrained using unlabeled data from both cell lines and patients. It is then fine-tuned using labeled data from cell lines and used for zero-shot testing on patients. The encoders and decoder in the autoencoder are three-layer neural network modules with dimensions (\emph{N}, 512), (512, 256), (256, 128) and (128, 256), (256, 512), (512, \emph{N}), respectively, and employ the rectified linear activation function. \emph{N} is the size of GEPs input, determined by tumor types and pretraining strategy. Details are described in the Supplementary materials. The classifier network used for fine-tuning consists of three-layer neural networks with dimensions (428, 64), (64, 32) using rectified linear activation, and (32, 1) using sigmoid activation.

\subsubsection*{The DSN model and its variants}

The DSN model and its variants also utilize an autoencoder backbone for unlabeled training. The key difference is that the DSN model incorporates two encoders with orthogonal embeddings. The underlying assumption is that GEPs consist of both drug-response-related signals and other confounding factors, which can be encoded by the shared and private encoders, respectively. Moreover, the distributional difference of the latent embeddings between the two domains (cell lines and patients) is minimized through maximum mean discrepancy (MMD) and adversarial loss. These losses can be applied either to the concatenation of private and shared embeddings or only to the shared embeddings, corresponding to the domain separation and reconstruction network (DSRN) and DSN models, respectively. In total, the DeepCRE model zoo encompasses eight models with various training losses. For mathematical details, please refer to the Supplementary materials.

\subsection*{The tumor-type adaptive pretraining strategy}

The tumor-type adaptive pretraining strategy is proposed to enhance the performance of DeepCRE models. Considering the significant pan-cancer heterogeneity in patient GEPs, not all patient data is suitable for pretraining a specific tumor type. tumor-type adaptive pretraining is considered more effective than the original all-data pretraining, particularly for adv-loss-based DeepCRE models (Fig. 2d). For instance, before zero-shot testing on drug response data for BRCA patients, the GEPs of both BRCA patients and all cell lines are utilized for unlabeled pretraining, followed by fine-tuning using cell line drug response data with the overlap drugs. Finally, this model is employed for zero-shot testing on the BRCA patient drug response dataset.

\noindent To demonstrate the impact of pretraining strategies on DeepCRE models, t-SNE plots are employed to visualize the distribution of embeddings for both cell lines and patients. The pretrained encoders from different strategies (test-pairwise or all-data) and methods (AE, DSN, DSN-mmd, DSN-adv) are saved and used to generate latent embeddings. These embeddings are then used for the t-SNE plot, where colors distinguish the sample types (Fig. 3d). The degree to which two colors mix together represents the alignment effect between the two domains. Two metrics, namely MMD and KL divergence, are employed to assess the alignment effect quantitatively.

\subsection*{Validation of DeepCRE for pharmaceutical value assessment}

To validate the potential of DeepCRE in clinical pharmaceutical value assessment, the DSN-adv DeepCRE model, trained with all-drug data, was employed to screen 233 small molecules for patients across 13 tumor types. The workflow of this screening process is illustrated in Fig. 4a. Firstly, the DSN-adv model generated a scoring matrix of drug response for patients (PDR) across 13 tumor types, such as COAD with 288 patients (Fi. 4b). Secondly, the top 5\% efficient drugs for each patient were filtered out to construct the efficient drugs for patients (EPD) pairs. The histogram plot of EPD pairs revealed that most drugs were inefficient, but a few drugs showed potential efficacy for over half of COAD patients (Fig. 4c and Supplementary Fig. 9). Thirdly, the top 10 drugs with the highest number of efficient patients were selected to form the efficient drugs for indications (EDI) list (Supplementary Table 9). The EDI lists across 13 tumor types were merged to generate the drug efficacy of indication (DEI) table. 

\noindent Additionally, five C-MDL models were utilized in this workflow to discover drug candidates for comparison. However, the P-SDL models were not applicable in this context as they require patient response data for every drug, which was unavailable.

\noindent DEI tables were utilized to verify drug efficacy within specific tumor types. These tables, specific to each of the 13 tumor types, contained information on the efficacy of 17 small molecules, which were considered drug candidates. To gather relevant records for these drug candidates, three databases were accessed: DrugBank~\cite{wishart2018drugbank}, ClinicalTrials~\cite{zarin2011clinicaltrials}, and Repurposing Hub~\cite{corsello2017drug}. Consequently, 10, 9, and 4 drug candidates had corresponding records in these databases, respectively. The information available in DrugBank and ClinicalTrials included tumor type specificity for the drug candidates. This information was then matched with the DEI tables using a heatmap visualization (Fig. 4e,g). Moreover, the ClinicalTrials and Repurposing Hub databases provided additional drug-related information, such as MoA, disease type, clinical phase, research status, etc. Sankey plots were employed to present the relationship between the drug candidates and these research items intuitively and quantitatively (Fig. 4f and Supplementary Fig. 10).

\noindent For a fair comparison with previous SOTA models, we defined "Qualified drugs" as the DEI drugs that met the criteria: i) no false negatives, which means excluding the drugs with any predicted-inefficient EDIs but in-clinical-test recorded EDIs, such as the drug candidate Rapamycin here even though there are some explanations; ii) at least one true positive, which means drugs need to have both predicted-efficient and in-clinical-test EDIs in at least one tumor type. This means that drugs with any predicted inefficient but in-clinical-test recorded EDIs were excluded, ensuring no false negatives. However, false positives were allowed in our analysis. This is because numerous drug-indication combinations have not been tested, and our scenario considers them "promising".

\subsection*{Identification of four types of drug sets}
DeepCRE were employed to generate the DRP score matrix for the patients of COAD and READ, two subtypes of colorectal cancer (CRC). This matrix facilitated the ranking of drug candidates based on their efficacy across patient populations, as illustrated in Supplementary Figure 13a,b. Additionally, we conducted a comprehensive search for clinical records of these drug candidates within the ClinicalTrials database. For clarity and ease of comparison, the evaluation outcomes of DeepCRE alongside traditional methodologies were depicted in an XY plot (Fig. 5c). Notably, the analysis delineated the drug candidates into four distinct sets (A, B, C, and D). Sets A and B comprised drugs favored by DeepCRE predictions but with limited clinical trial evidence, whereas Sets C and D included drugs that were either approved or had substantial clinical trial involvement but were underestimated by DeepCRE predictions.

\subsection*{Experimental assays for the identified drug candidates}
To evaluate the efficacy of the identified drug candidates, we employed the In-SMARchip methodology. A total of 16 drug candidates were tested across four concentration gradients (0.2, 2, 20, and 200 uM), with each assay conducted in triplicate. These experiments were performed on eight CRC organoid models. The methodological details, including the experimental setup and analysis, are consistent with those described in our previous work~\cite{hu2021lung, wu2022grouped, wang2018nanoliter}.

\subsection*{Docking analysis of small molecules with specific target}
The interaction between Sorafenib and the ATP-binding cassette sub-family G member 2 (ABCG2) was meticulously analyzed using Autodock Vina version 1.2.2~\cite{seeliger2010ligand}. This analysis aimed to elucidate the binding poses and molecular interactions critical for the drug's mechanism of action. These interactions were visualized through the PyMOL software~\cite{seeliger2010ligand}, providing a detailed representation of Sorafenib's binding efficiency and orientation to ABCG2.

\subsection*{Bioinformatics analysis of the pharmacological profile for drug candidates}
To discern the pharmacological distinctions between GSK1059615 and other inhibitors targeting the PI3K/mTOR pathway, we extracted perturbation data from the Connectivity Map (CMAP) L1000 database~\cite{subramanian2017next}. This dataset was curated to include GSK1059615, WYE-125132 (a comparator PI3K/mTOR inhibitor), and DMSO (as control). Initially, a Principal Component Analysis (PCA) was conducted on the 2,000 most variable genes, followed by a two-dimensional Uniform Manifold Approximation and Projection (Umap) analysis based on the top 50 principal components~\cite{hao2021integrated}. Differential expression analysis further identified representative genes, which were visualized on the Umap plot (Fig. 6e). Subsequently, pathway enrichment analysis was performed~\cite{yu2012clusterprofiler}, and the top five up-regulated and down-regulated pathways, determined by adjusted p-values, were illustrated in Fig. 6f, providing insight into the molecular underpinnings and potential therapeutic implications of GSK1059615 compared to other PI3K/mTOR inhibitors.

\subsection*{Code availability}
The source code is available at \url{https://github.com/wuys13/Multi-Drug-Transfer-Learning.git}.

\subsection*{Data availability}
DeepCRE dataset, intermediate results, and the final drug response predictions can be found at \url{https://zenodo.org/record/8021167}.

\newpage
\bibliography{bibliography}
\bibliographystyle{naturemag}

\newpage
\section*{Acknowledgements}
This research was substantially sponsored by the research projects (Grant No. 2022YFF1203000) supported by the National Key Research and Development Program of China and was substantially supported by the Beijing Academy of Artificial Intelligence (BAAI).


\section*{Declaration of Interests}
The authors declare no competing interests.



\end{document}


\maketitle

\beginsupplement

\newpage
\section*{Supplementary Figures}

\begin{figure}[h]
	\begin{center}
		\includegraphics[width=\textwidth]{supfig1}
	\end{center}
	\caption{\textbf{\lipsum[30][1]} \lipsum[30][2]}
	\label{sup.fig:my_sup_plot1}
\end{figure}

\clearpage
\begin{figure}[h]
	\begin{center}
		\includegraphics[width=\textwidth]{supfig2}
	\end{center}
	\caption{\textbf{\lipsum[30][3]} \lipsum[30][4]}
	\label{sup.fig:my_sup_plot2}
\end{figure}

\newpage
\section*{Supplementary Methods}

\subsection*{\lipsum[45][1]}
\lipsum[40][1]
\cite{pearson1901liii}
\lipsum[40][2]
\ref{fig:my_plot1} and \ref{fig:my_plot2}.
\lipsum[40][3-100]

\subsection*{\lipsum[45][2]}
\lipsum[41][1-3]
\cite{chopra2005learning}
\lipsum[41][4-100]
\ref{sup.fig:my_sup_plot1}

\lipsum[51][1]
\cite{hermans2017defense}
\lipsum[51][2-100]
\ref{sup.fig:my_sup_plot2} and \ref{sup.fig:my_sup_plot3}

\begin{figure}[h]
	\makebox[\textwidth]{\includegraphics[width=0.5\textwidth]{supfig3}}
	\centering
	\caption{\textbf{\lipsum[30][5]} \lipsum[31][1-3]}
	\label{sup.fig:my_sup_plot3}
\end{figure}

\clearpage
\printbibliography[title=Supplementary References]